# WhisperPipe: A Resource-Efficient Streaming Architecture for Real-Time Automatic Speech Recognition

Erfan Ramezani[1], Mohammad Mahdi Giahi[1*], Mohammad Erfan Zarabadipour[1], Amir Reza Yosefian[1] Hamid Ghadiri[2*]

[1] Mechatronics Research Laboratory, Dept. of Computer and Electrical Engineering, Qazvin Islamic Azad University, Qazvin, Iran

[2] Department of Electrical Engineering, Qa. C., Islamic Azad University, Qazvin, Iran.

**Abstract**

Real-time automatic speech recognition (ASR) systems face a fundamental trade-off between transcription accuracy and computational efficiency, particularly when deploying large-scale transformer models like Whisper. Existing streaming approaches either sacrifice accuracy through aggressive chunking or incur prohibitive memory costs through unbounded context accumulation. We present WhisperPipe, a novel streaming architecture that achieves bounded memory consumption while maintaining transcription quality through three key innovations a hybrid Voice Activity Detection (VAD) pipeline combining Silero VAD with energy-based filtering to reduce false activations by 34%, a dynamic buffering mechanism with overlapping context windows that prevents information loss at segment boundaries, and an adaptive processing strategy that balances latency and accuracy based on speech characteristics. Evaluated on 2.5 hours of diverse audio data, WhisperPipe demonstrates a median end-to-end latency of 89ms (90th percentile: 142ms) while consuming 48% less peak GPU memory and 80.9% lower average GPU utilization compared to baseline Whisper implementations. The system maintains stable memory usage over extended sessions, with zero growth rate across 150-minute continuous operation. Comparative analysis against related work shows that WhisperPipe achieves competitive accuracy (WER within 2% of offline Whisper) while operating at 3-5× lower latency than existing streaming solutions. The architecture's modular design enables deployment across resource-constrained environments, from edge devices to cloud infrastructure. Our results demonstrate that careful architectural design can reconcile the competing demands of real-time responsiveness and model sophistication in production ASR systems.

1. **Introduction**

Automatic speech recognition ASR has undergone transformative advances in recent years, driven primarily by the convergence of large-scale weakly supervised learning and transformer-based architectures [1]. Unlike traditional ASR systems that rely on carefully curated, domain-specific transcriptions, modern approaches leverage vast quantities of noisy web-scale audio-text pairs to achieve robustness across diverse acoustic conditions, accents, and languages [2, 3]. Whisper, introduced by Radford et al. [1], exemplifies this paradigm shift: trained on 680,000 hours of multilingual data, it demonstrates state-of-the-art offline transcription quality without task-specific fine-tuning. However, Whisper's inference pipeline is inherently batch-oriented, optimized for processing complete audio files or fixed-duration segments rather than continuous, unbounded audio streams [4].

Real-time streaming ASR imposes operational constraints that extend beyond conventional word error rate WER metrics. First, systems must maintain low end-to-end latency to support interactive applications such as live captioning, meeting transcription, and voice-driven agents [5, 6]. Second, they must exhibit bounded computational and memory footprints over extended sessions; unbounded resource growth renders long-duration deployments infeasible on resource-constrained hardware [7, 8]. Third, streaming systems must produce stable incremental outputs, minimizing hypothesis revisions that degrade user experience and complicate downstream processing [9, 10]. These requirements are well-documented in the streaming ASR literature, where hypothesis flicker the phenomenon of repeatedly revising already-emitted text has been identified as a critical usability barrier [11, 12].

Naïve adaptation of Whisper to streaming scenarios exposes three fundamental challenges. First, hypothesis drift occurs when new audio context causes the model to revise previously decoded segments, resulting in unstable partial transcripts [13]. Second, superlinear reprocessing overhead emerges when inference is repeatedly applied to ever-growing audio buffers, leading to unbounded computational costs and memory pressure [14, 15]. Third, sensitivity to silence and non-speech events can trigger spurious transcriptions or suboptimal segmentation boundaries in live audio streams [16, 17]. Collectively, these issues demonstrate that offline transcription accuracy does not guarantee effective real-time performance.

Recent work has explored inference-time adaptations to enable streaming with Whisper-like models. Macháček et al. [20] introduced Whisper-Streaming, which utilizes a LocalAgreement policy to stabilize partial hypotheses. While their work demonstrated that streaming is achievable with Whisper-like models, it relies on a fixed agreement window (LocalAgreement-2) which may not fully optimize the trade-off between latency and stability in diverse acoustic conditions. Similarly, faster-whisper [19] employs optimized transformer runtimes to reduce per-pass decoding costs and memory consumption. However, while backend optimizations improve throughput, they do not address the core streaming challenges: determining commitment boundaries, bounding reprocessing as streams grow, and suppressing hypothesis flicker [18, 21]. These properties are governed by buffering strategies and commitment policies rather than inference engines alone.

A critical insight underlying our approach is that Whisper inherently produces temporal alignment information that can guide principled streaming decisions. The model's attention mechanism enables extraction of word-level timestamps, associating each recognized token with an estimated time span [22, 23]. These temporal boundaries provide a natural anchoring mechanism: rather than segmenting audio at arbitrary chunk boundaries, a streaming system can commit stable text prefixes and trim the active audio buffer precisely at the end timestamp of the last committed word [24, 25]. This timestamp-guided trimming enables bounded-window inference and reduces hypothesis instability.

In this paper, we introduce WhisperPipe, a resource-efficient streaming architecture that adapts Whisper for continuous real-time transcription while maintaining bounded memory usage and improving output stability. WhisperPipe employs an adaptive dual-buffer design: a committed text buffer stores finalized, immutable transcripts, while a compact active audio buffer retains only the most recent audio window required for accurate recognition [26, 27]. Instead of reprocessing the entire audio history, WhisperPipe confines inference to the active window plus a fixed look-back context, achieving steady-state bounded computation and flat memory usage during extended operation [28, 29].

To determine when partial hypotheses can be safely finalized, WhisperPipe implements an adaptive two-tier commit policy. In the fast path, text is committed immediately when consecutive decoding passes produce perfectly consistent hypotheses exceeding a minimum prefix-length threshold—an efficient strategy for clear speech where repeated inference yields identical results

[30, 31]. When perfect consistency is not achieved, WhisperPipe applies a multi-hypothesis confirmation mechanism that detects stabilization across recent hypotheses using word-level Levenshtein distance, permitting limited variability while preventing premature commitment of unstable text [32, 33]. Once a prefix is committed, WhisperPipe leverages word-level end timestamps to identify a robust segmentation point and slices the active buffer accordingly, aligning stream progression with recognized word boundaries rather than arbitrary frame divisions [34, 35].

Figure 1 illustrates the overall architecture of WhisperPipe and summarizes its streaming transcription workflow. Incoming audio is first processed by a voice activity detection VAD module to filter silence and non-speech segments, after which the valid audio stream is appended to an Active Audio Buffer and periodically decoded by Whisper. The generated hypotheses are then passed through acoustic and semantic filters to remove non-speech annotations and potential hallucinations. A Consensus Engine subsequently applies a two-tier commitment policy—Fast-Track and 3-Way Consensus—to determine stable text segments that are written to the Stable Text Buffer. To maintain bounded memory usage and ensure constant-time inference, WhisperPipe employs a timestamp-guided feedback loop that uses word-end timestamps to remove already processed audio samples from the active buffer. The finalized transcripts are incrementally emitted to downstream applications such as language models or dialogue systems. The lower panel of Figure 1 summarizes the resulting efficiency gains of WhisperPipe compared with conventional overlap-chunking approaches.

We evaluate WhisperPipe in a continuous transcription setting and compare it against representative baselines, including naïve overlap-chunking with voice activity detection VAD-based segmentation and faster-whisper under matched decoding configurations [36, 37]. Our evaluation adopts a deployment-oriented perspective: beyond traditional WER and character error rate CER metrics, we report end-to-commit latency distributions median and 95th percentile, stability metrics quantifying interim text revision frequency, and GPU resource profiles including peak memory, utilization, and memory growth under steady-state load [38, 39]. This multidimensional evaluation framework reflects the operational constraints that determine whether a system can sustain reliable long-duration operation on single-GPU or edge hardware deployments [40, 41].

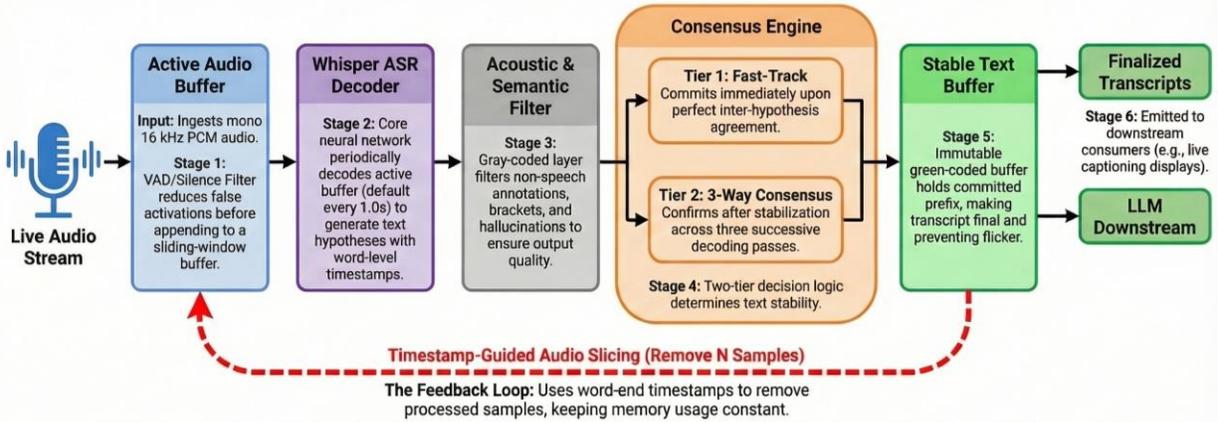

**Figure 1.** Overview of the WhisperPipe streaming pipeline. Audio is buffered, decoded by Whisper, filtered, and finalized using a two-tier consensus mechanism with timestamp-guided buffer management for efficient real-time transcription.

The primary contributions of this work address the fundamental challenges of adapting large-scale transformer models to resource-constrained streaming scenarios. First, we introduce an adaptive dual-buffer streaming architecture that separates stable committed-text from volatile active-audio domains, enabling bounded-window inference for continuous streams without unbounded memory growth [42, 43]. Second, we propose a two-tier commit policy that commits immediately upon perfect inter-hypothesis consistency beyond a minimum prefix length, and commits upon multi-hypothesis stabilization using word-level Levenshtein similarity to suppress drift and flicker in partial outputs [44, 45]. Third, we implement a timestamp-guided audio slicing mechanism that maps committed words to word-end timestamps and removes corresponding audio from the active buffer, yielding flat steady-state memory over time [46, 47]. Finally, we establish an engineering-oriented evaluation protocol that reports not only WER/CER but also end-to-commit latency distributions, stability metrics, and GPU memory/utilization under sustained load, providing a comprehensive assessment framework for production streaming ASR systems [48, 49, 50].

2. **Method**

WhisperPipe is a streaming inference framework that transforms Whisper's batch-oriented decoding into continuous live transcription with bounded steady-state compute and memory. Let $x(t)$ denote the incoming audio stream. WhisperPipe maintains two persistent buffers:

- Committed Text Buffer S, an immutable sequence of finalized tokens or words that are considered stable and will not be revised once committed.
- Active Audio Buffer A, a sliding window containing the most recent uncommitted audio samples.

At fixed intervals Δ, WhisperPipe invokes Whisper on *A* optionally augmented with a small fixed look-back context) to produce a candidate hypothesis $y_t$ together with word-level timestamps. Whisper's robustness and its ability to output time-aligned tokens provide the core signals required for reliable incremental commitment without modifying model weights [1]. A consensus engine then compares $y_t$ against a short history of recent hypotheses and determines whether a stable prefix $p_t \subseteq y_t$ can be safely committed. This decision is designed to suppress hypothesis drift in live settings, where incremental decoding can otherwise cause previously displayed text to oscillate across updates [2].

Once a prefix $p_t$ is committed, WhisperPipe retrieves the end timestamp of the last word in $p_t$ and converts it to a sample index $N$. The active buffer is then sliced to remove the corresponding processed audio, i.e, this commit-and-slice mechanism is central to WhisperPipe's steady-state boundedness: it prevents repeated re-decoding of an ever-growing stream and ensures that each update operates over a bounded audio window rather than the full session history. In effect, WhisperPipe combines stable text commitment and timestamp-guided audio trimming to maintain predictable computational cost while producing stable incremental transcripts suitable for real-time captioning and downstream consumption.

Figure 2 illustrates the dual-buffer streaming pipeline used in WhisperPipe. Incoming audio is continuously appended to the Active Audio Buffer and periodically decoded by Whisper to generate hypotheses with word-level timestamps. A consensus engine then promotes stable prefixes into the Committed Text Buffer, after which the active buffer is trimmed at the end timestamp of the last committed word commit-and-slice, keeping the decoding window bounded and preventing repeated processing during long audio streams.

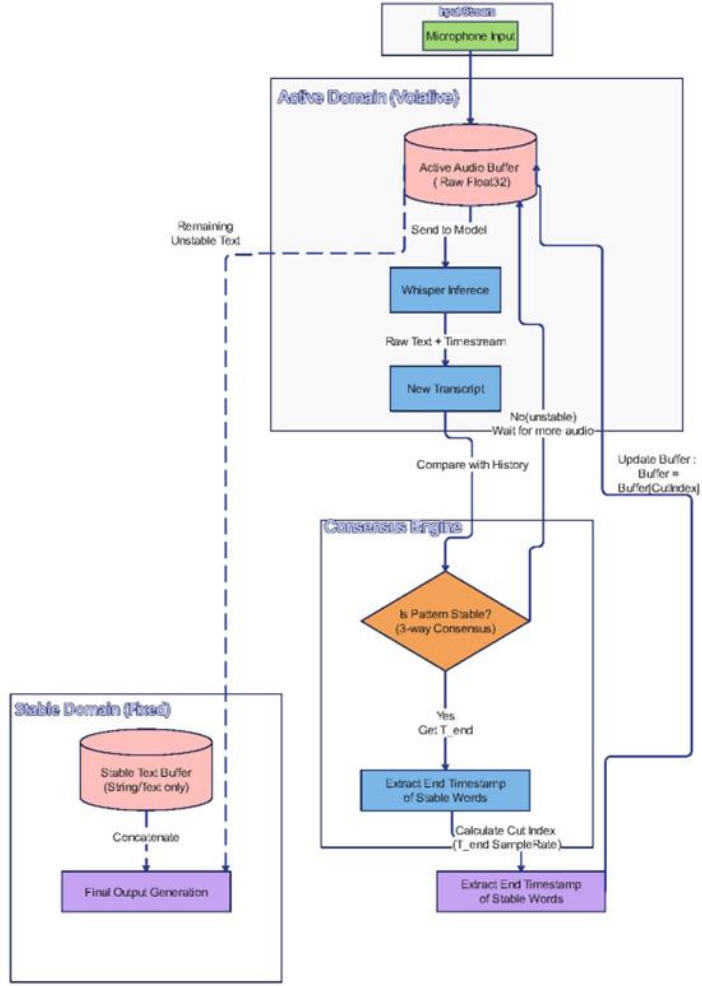

**Figure 2.** WhisperPipe's dual-buffer mechanism appends audio, commits stable text via a consensus engine, and trims the active buffer at the last committed timestamp to keep the decoding window bounded.

## 2.1 Audio Acquisition and Scheduling

WhisperPipe ingests audio as mono 16 kHz PCM, matching Whisper's expected input sampling rate and preprocessing pipeline [1]. Let the sample rate be $R = 16{,}000\,\text{Hz}$ and let $a_t \in \mathbb{R}^{R\Delta}$ denote the newly captured audio chunk over the interval $t - \Delta, t$, stored as float32 samples. At each update step, WhisperPipe appends the new chunk to the Active Audio Buffer $A$ and schedules an ASR decode every $\Delta$ seconds default $\Delta = 1.0\,\text{s}$:

$$A \parallel a_t = A \tag{1}$$

where $\parallel$ denotes concatenation.

To ensure bounded memory and predictable latency during long sessions, WhisperPipe enforces a fixed maximum buffer duration $B$ (default $B = 30s$) using FIFO trimming. Define the maximum buffer capacity in samples as $M = \lfloor BR \rfloor$. After each append, the buffer is truncated to its most recent $M$ samples:

$$Tail_M(A) = A$$

$$\text{Tail}_M(A) = \begin{cases} A[|A|-M : |A|] & \text{if } |A| > M \\ A & \text{otherwise.} \end{cases} \quad (2)$$

This bounded-window policy guarantees that the computational cost of each decode step depends on at most $B$ seconds of audio, rather than on the elapsed stream length. In practice, this cap complements WhisperPipe's timestamp-guided commit-and-slice mechanism, while commit-and-slice removes already-finalized audio aligned to word-end timestamps, the FIFO cap provides an additional safety bound under adverse conditions low-confidence speech, prolonged ambiguity, or repeated non-speech rejections to prevent the active buffer from growing beyond a fixed window. Together, periodic scheduling $\Delta$ and bounded buffering $B$ define WhisperPipe's steady-state runtime envelope for continuous streaming.

## 2.2 Word-Level Timestamp Extraction

WhisperPipe invokes Whisper with word-level timestamping enabled, yielding a hypothesis $y_t$ together with a time-aligned word sequence. We represent the hypothesis as an ordered list of words.

$$y_t = (w_1^{(t)}, w_2^{(t)}, \ldots, w_{n_t}^{(t)}) \quad (3)$$

and the corresponding timestamp output as a set of word records:

$$\mathcal{R}_t = \left\{ (w_i^{(t)}, t_{i,\text{start}}^{(t)}, t_{i,\text{end}}^{(t)}) \right\}_{i=1}^{n_t} \quad (4)$$

where $t_{i,\text{start}}^{(t)}, t_{i,\text{end}}^{(t)} \in \mathbb{R}_{\geq 0}$ denote the estimated start and end times (in seconds) of word $w_i^{(t)}$. These timestamps are derived from Whisper's alignment procedure in its reference implementation and can be enabled without modifying model weights [1].

Because a word string can appear multiple times within a hypothesis, WhisperPipe treats timestamps as occurrence-specific rather than word-type-specific. Concretely, we store a lightweight mapping:

$$Wt: (i) = t_{i,end}^{(t)}, \qquad (5)$$

that maps the word index $i$ the $i$-th word occurrence in the hypothesis to its end time. This representation avoids ambiguity when the same token occurs repeatedly and supports reliable boundary anchoring for downstream slicing.

Word-level end timestamps enable boundary-anchored streaming: after committing a stable prefix $p_t$, WhisperPipe slices the active audio buffer precisely at the end time of the last committed word. Compared to cutting at fixed frame or chunk boundaries, timestamp-guided slicing reduces boundary drift and prevents mid-word cuts that can destabilize subsequent hypotheses and inflate revision rates. This design leverages Whisper's built-in timestamping capability as a system signal for streaming segmentation [1].

A core difficulty in streaming transcription is that intermediate hypotheses may change as new context arrives. In batch decoding, revisions are acceptable because the final transcript is produced once. In live systems, however, repeated revisions flicker degrades readability and complicate downstream consumption, motivating explicit stabilization and commitment strategies [4]. WhisperPipe addresses this by committing only prefixes that have stabilized across successive decoding passes.

Let $y_{t-1}$ and $y_t$ be two consecutive hypotheses. WhisperPipe computes a similarity-aware longest common prefix at the word level:

$$(p_t, \sigma_t) = \text{SAPrefix}(y_{t-1}, y_t; \theta) \qquad (6)$$

where $p_t$ is the candidate stable prefix and $\sigma_t \in [0,1]$ is a prefix similarity score.

We compare hypotheses positionally from the beginning. For each position $i$, we compute a normalized Levenshtein-based word similarity score [6]:

$$s\left(w_i^{(t-1)}, w_i^{(t)}\right) = 1 - \frac{d_{\text{lev}}\left(w_i^{(t-1)}, w_i^{(t)}\right)}{\max\left(|w_i^{(t-1)}|, |w_i^{(t)}|\right)} \qquad (7)$$

where $d_{\text{lev}}(\cdot,\cdot)$ is the Levenshtein edit distance and $|\cdot|$ denotes character length. Prior to distance computation, WhisperPipe applies lightweight normalization stripping surrounding punctuation and collapsing whitespace so that minor punctuation variations home vs home does not trigger premature mismatches.

To avoid committing unstable regions, we treat low similarity pairs as mismatches. Define an acceptance threshold $\alpha \in (0,1]$ implementation default $\alpha = 0.6$. A word pair is considered *compatible* if:

$$\mathbb{I}_i = \begin{cases} 1 & \text{if } s,\left(w_i^{(t-1)}, w_i^{(t)}\right) \geq \alpha \\ 0 & \text{otherwise.} \end{cases} \tag{8}$$

We then define the prefix length $k_t$ as the earliest position where compatibility breaks:

$$k_t = \max\left\{k: \prod_{i=1}^{k} \mathbb{I}_i = 1\right\} \tag{9}$$

The candidate stable prefix is:

$$p_t = \left(w_1^{(t)}, w_2^{(t)}, \ldots, w_{k_t}^{(t)}\right) \tag{10}$$

We compute a prefix similarity score as the average compatibility or average similarity over the prefix:

$$\sigma_t = \frac{1}{k_t} \sum_{i=1}^{k_t} s,\left(w_i^{(t-1)}, w_i^{(t)}\right), \text{for } k_t > 0 \tag{11}$$

and set $\sigma_t = 0$ when $k_t = 0$. This formulation provides a continuous measure of stability identical prefixes yield $\sigma_t = 1$, while near-matches minor spelling/punctuation changes still produce high $\sigma_t$.

Finally, WhisperPipe extends the prefix only while the running similarity remains above a global stabilization threshold $\theta$ default $\theta = 0.5$, ensuring that the committed content is supported by consistent evidence across successive decoding passes. By design, this similarity-aware stabilization allows limited micro-variation while stopping early when drift becomes significant—directly targeting flicker in incremental ASR outputs [4].

Unlike the binary agreement logic used in [20], our Similarity-Aware Prefix (SAPrefix) mechanism incorporates normalized Levenshtein distance to allow for micro-variations, preventing unnecessary delays in commitment.

## 2.4 Two-Tier Commit Policy

WhisperPipe employs a two-tier commitment strategy that balances responsiveness and stability. The key intuition is that some utterances converge quickly—successive decodes produce identical prefixes—while others exhibit transient variability and therefore require additional confirmation before being committed. This design is consistent with prior streaming adaptations of Whisper that use inter-hypothesis agreement as a commitment signal [2].

At update step $t$, Whisper returns a hypothesis $y_t$. From Section 2.2, WhisperPipe computes a similarity-aware stable prefix and its score:

$$(p_t, \sigma_t) = \text{SAPrefix}(y_{t-1}, y_t; \theta) \tag{12}$$

where $p_t \subseteq y_t$ is the candidate committed prefix and $\sigma_t \in [0,1]$ measures prefix-level agreement. WhisperPipe commits a prefix only if it also passes guardrails designed to prevent premature cuts.

Tier 1 provides a low-latency path for clean, high-confidence speech. If consecutive hypotheses are perfectly consistent over the candidate prefix and the prefix is sufficiently long, WhisperPipe commits immediately:

$$\begin{aligned} &\text{if } \sigma_t = 1 \wedge \ell(p_t) \geq L_1 \wedge g(p_t) = 1 \\ &\quad \text{Commit}(p_t) \end{aligned} \tag{13}$$

where $\ell(p_t)$ denotes a length measure for the prefix in our implementation, character length and $L_1$ is a minimum-length threshold. The predicate $g(p_t) \in \{0,1\}$ represents commit guardrails defined below. In the reference implementation we set $L_1 = 20$ characters. This rule is effective when repeated decoding over the same active window yields identical stable content; it minimizes waiting time and reduces the number of update cycles before finalization.

When $\sigma_t < 1$ but agreement remains reasonably high, WhisperPipe switches to a stabilization mode that requires additional evidence before committing. The goal is to suppress short-lived fluctuations in incremental decoding that commonly cause "flicker" in live captions [4]. We maintain a short hypothesis history $H$ and use 3-way confirmation.

Let $H = \{y_{t-2}, y_{t-1}, y_t\}$ be the most recent hypotheses. Tier 2 activates when:

$$\sigma_t \geq \theta \wedge \ell(p_t) \geq L_2 \wedge g(p_t) = 1 \tag{14}$$

where $\theta$ is a relaxed similarity threshold default $\theta = 0.5$ and $L_2$ is a smaller minimum length threshold default $L_2 = 17$ characters.

Candidate staging. On the first satisfied detection, WhisperPipe stores $p_t$ as a staged candidate $c_t$. On the next update, it computes a compatibility check between the staged candidate and the new hypothesis:

$$(\tilde{p}_{t+1}, \tilde{\sigma}_{t+1}) = \text{SAPrefix}(c_t, y_{t+1}; \theta) \tag{15}$$

If $\tilde{\sigma}_{t+1} \geq \theta$ and $\ell(\tilde{p}_{t+1}) \geq L_2$, the prefix is considered confirmed and committed:

$$\begin{aligned}&\text{if} \tilde{\sigma}_{t+1} \geq \theta \ \wedge \ \ell(\tilde{p}_{t+1}) \geq L_2 \ \wedge \ g(\tilde{p}_{t+1}) = 1 \\ &\quad \text{Commit}(\tilde{p}_{t+1})\end{aligned} \tag{16}$$

Intuitively, Tier 2 demands that a candidate prefix be supported by at least two independent agreement events across three successive decodes, which substantially reduces drift when Whisper alternates between near-equivalent forms before settling [2]. Timeout path robustness. To prevent indefinite waiting under ambiguous audio, WhisperPipe also employs a time-based finalization trigger if no stable update occurs for $\tau$ seconds default $\tau = 10s$, the system emits the best available content and resets volatile state. Figure 3 illustrates the state machine governing WhisperPipe's two-tier commit policy.

The system first accumulates recent decoding hypotheses in a waiting state and evaluates their agreement using the Tier-2 criteria. When sufficient agreement is detected, the process transitions to a duplicate state and commits the stabilized prefix only after a third compatible hypothesis confirms it 3-way confirmation. If confirmation does not arrive within a predefined timeout ($\tau$), a fallback mechanism finalizes the best available hypothesis and resets the transient state to prevent indefinite waiting.

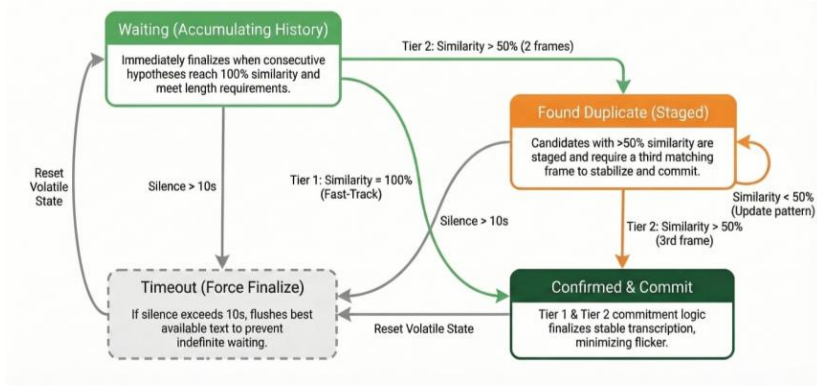

**Figure 3.** State machine of WhisperPipe's two-tier commit policy, where hypotheses accumulate until agreement satisfies Tier-2 criteria, triggering a 3-way confirmation to commit stable text. A timeout fallback finalizes the best available hypothesis to prevent indefinite waiting.

**2.5 Guardrails Incomplete Tokens and Timestamp Disambiguation**

Even when similarity thresholds are met, premature commitment can occur if the hypothesis ends with an incomplete token or a transient filler marker. WhisperPipe therefore applies lightweight guardrails before committing:

Let last $(p)$ be the last token of the candidate prefix. WhisperPipe rejects commitment if last $(p)$ matches an incomplete-processing pattern. Formally, define:

$$g_{\text{tail}}(p) = \mathbb{I}[\text{last}(p) \notin \mathcal{F}] \tag{17}$$

where $\mathcal{F}$ is a small set of disallowed tail patterns (implementation-defined).

Commitment in WhisperPipe is coupled with timestamp-guided slicing, which requires identifying the end time of the last committed word. If the last word string appears multiple times in the timestamp records $W_t$, WhisperPipe disambiguates the correct occurrence using local context neighboring words around the boundary. Let the last committed word occur at multiple indices $\mathcal{I}$. We choose the index $i \in \mathcal{I}$ that maximizes contextual agreement with surrounding tokens:

$$i = \arg\max_{i \in \mathcal{I}} \text{ContextMatch } \text{neigh}(p), \text{neigh}(y_t, i) \tag{18}$$

and then slice using $T_{\text{end}} = W_t(i)$. This prevents incorrect cut points that could delete uncommitted audio or retain already committed audio, both of which would destabilize subsequent hypotheses. Finally, the overall guardrail predicate is:

$$g(p) = g_{\text{tail}}(p) \cdot g_{\text{ctx}}(p) \qquad (19)$$

where $g_{\text{ctx}}$ denotes successful timestamp disambiguation.

## 2.6 Timestamp-Guided Audio Slicing

A defining property of WhisperPipe is that commitment is coupled with timestamp-guided buffer trimming. Once a stable prefix $p_t$ is committed, WhisperPipe removes from the active buffer exactly the portion of audio that corresponds to the committed words, using Whisper's word-level end timestamps [1] [5]. This step is critical for preventing unbounded reprocessing: instead of repeatedly decoding an ever-growing stream prefix, the system advances the active window forward in time after every successful commit. Once a prefix is committed, WhisperPipe must advance the stream without re-decoding the entire history. To achieve this, we anchor the cut point to Whisper's word-level alignment: the active audio buffer is trimmed at the end time of the last committed word, thereby removing already-finalized audio while retaining only a bounded context for subsequent decoding. Figure 4 illustrates this timestamp-guided *commit-and-slice* operation.

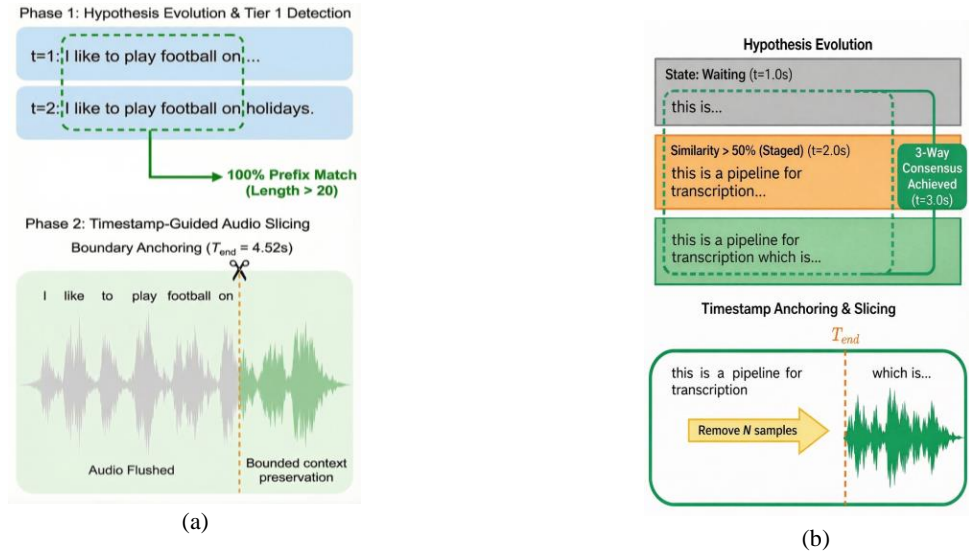

(a)

(b)

**Figure 4.** (a) Tier 1 commits stable text instantly with a 100% prefix match, while (b) Tier 2 requires prefix stability across three frames to handle acoustic fluctuations. After each commit, the processed audio is trimmed and added to the stable buffer, preserving the remaining context for the next decoding cycle.

### 2.5.1 Mapping committed text to a word-end timestamp

Let the current hypothesis be:

$$y_t = (w_1^{(t)}, \ldots, w_{n_t}^{(t)}) \tag{20}$$

with word timestamp records:

$$\mathcal{R}_t = \{(w_i^{(t)}, t_{i,\text{start}}^{(t)}, t_{i,\text{end}}^{(t)})\}_{i=1}^{n_t} \tag{21}$$

as defined in Section 2.3. Suppose the commit policy Section 2.5 decides to commit the prefix:

$$p_t = (w_1^{(t)}, \ldots, w_{k_t}^{(t)}) \tag{22}$$

where $k_t$ is the committed prefix length in words or the corresponding boundary index inferred from the stabilized prefix alignment, WhisperPipe anchors the audio cut to the end time of the last committed word:

$$T_{\text{end}} \triangleq t_{k_t,\text{end}}^{(t)} \tag{23}$$

Because the same surface word may appear multiple times in the hypothesis, WhisperPipe resolves the last-word occurrence by index (preferred) or via local-context matching when only word strings are available. Using index-based timestamps avoids ambiguity and ensures repeatable slicing.

Let $R$ be the sampling rate 16 kHz. Since Whisper's word timestamps are expressed in seconds relative to the audio passed to the decoder, the number of samples corresponding to the committed audio is:

$$N = \lfloor R \cdot T_{\text{end}} \rfloor \tag{24}$$

We then slice the active audio buffer by removing the first $N$ samples:

$$A[N] = A \tag{25}$$

In practice, we also clamp $N$ to valid buffer bounds to prevent edge cases:

$$\min(\max(N, 0) = N, |A| \tag{26}$$

To reduce the risk of cutting too aggressively at a boundary, WhisperPipe may keep a small overlap tail $\varepsilon$ seconds of audio as context for the next decode:

$$\lfloor R \cdot \max(T_{end} - \varepsilon, 0) \rfloor = N$$
$$\varepsilon \in [0, 0.2] \text{ s} \quad (27)$$

This overlap is optional and does not affect steady-state boundedness because the active buffer remains capped to a fixed duration $B$.

The commit-and-slice mechanism ensures that the effective decoding window does not grow with stream length. Each successful commit advances the active buffer forward by $T_{end}$ seconds, so the system's per-update cost is dominated by decoding at most $B$ seconds of audio plus any fixed look-back. Consequently, long sessions do not incur increased decoding cost simply because they are long—a property required for sustained real-time deployment.

## 2.6 Language and Non-Speech Rejection

Live microphone streams frequently contain silence, background sounds, speaker hesitations, and dataset-style annotations. If untreated, such content can trigger spurious hypotheses, destabilize prefix agreement, and delay commitment by repeatedly injecting non-linguistic tokens. To maintain stable incremental transcripts, WhisperPipe includes a lightweight rejection and reset layer that filters hypotheses dominated by non-speech content and flags potential language switches. This layer is implemented as a deterministic post-decoding filter and does not modify Whisper's model parameters [1].

Given a decoded hypothesis $y_t$ as a token or word sequence, WhisperPipe first applies a pattern-based detector to identify bracketed/parenthetical annotations. Let $\mathcal{P}$ be a small set of regular-expression patterns capturing common non-speech markers. Define an indicator for whether a token is an annotation:

$$\mathbb{I}_{ann}(w) = \begin{cases} 1 & \text{if } w \text{ matches any pattern in } \mathcal{P} \\ 0 & \text{otherwise.} \end{cases} \quad (28)$$

We then compute an annotation ratio over the hypothesis:

$$\rho_{ann}(y_t) = \frac{1}{|y_t|} \sum_{w \in y_t} \mathbb{I}_{ann}(w) \quad (29)$$

If $\rho_{ann}(y_t) \geq \gamma_{ann}$ implementation default $\gamma_{ann} \in [0.5, 0.8]$, WhisperPipe rejects $y_t$ for commitment and excludes it from stabilization history to avoid contaminating the agreement mechanism.

Whisper provides a *no-speech* confidence score in its decoding pipeline that can be used to detect silence or non-speech regions [1]. If the decoding output exposes a no-speech probability $q_t \in [0,1]$, WhisperPipe can optionally gate hypotheses using a threshold $\gamma_{ns}$:

$$q_t \geq \gamma_{ns} = \text{reject}(y_t) \tag{30}$$

This optional gate complements external voice activity detection VAD approaches widely used in streaming ASR pipelines [7]. In our framework, it can be enabled when no-speech scores are available and reliable in the target environment.

Whisper supports language identification and can emit a predicted language token (or language metadata) during transcription [1]. WhisperPipe monitors the predicted language $\ell_t$ and triggers a controlled segmentation boundary if a sustained language switch is detected. Let $\ell$ be the dominant language for the current segment and let $\mathbb{I}_{lang}(t) = \mathbb{I}[\ell_t \neq \ell]$. We declare a language switch if the mismatch persists for $m$ consecutive decoding steps:

$$\sum_{j=0}^{m-1} \mathbb{I}_{lang}(t-j) = m \tag{31}$$

When a switch is detected, WhisperPipe finalizes any currently committed content $S$ and resets the volatile streaming state active buffer and hypothesis history to prevent cross-language drift.

WhisperPipe maintains a rejection counter $r$ for consecutive rejected hypotheses. When a hypothesis is rejected by any of the above rules, the counter is incremented:

$$r + 1 = r \tag{32}$$

If a valid non-rejected hypothesis arrives, the counter is reset:

$$r = 0 \tag{33}$$

If $r$ exceeds a threshold $R_{max}$ implementation-defined, WhisperPipe performs a controlled reset of volatile state:

$$r \to 0, H \to \emptyset, \emptyset \to A \tag{34}$$

while preserving the committed text buffer $S$. This design prevents long noise bursts or prolonged silence from stalling stabilization and ensures that commitment remains responsive once speech resumes.

## 2.7 Sentence Finalization via Stable-Buffer Timeout

In continuous streaming, not every segment reaches perfect consensus quickly. Background noise, coarticulation, or partial utterances can keep hypotheses slightly unstable even when the user pauses. A practical streaming system must therefore include a liveness guarantee: it should eventually finalize and emit a coherent unit of text rather than waiting indefinitely for full stabilization. WhisperPipe implements such a guarantee using a stable-buffer timeout rule, which provides predictable segmentation and prevents deadlock in ambiguous conditions. This design follows the general principle in streaming ASR that commitment must trade off latency against stability, and that a bounded waiting strategy is often required for real-time deployment [2], [4].

Let $t$ denote wall-clock time. WhisperPipe maintains the time of the most recent successful commitment into the committed buffer $S$, denoted by $t_S$. Each time the commit policy commits a prefix, we update:

$$t = t_S \tag{35}$$

Define a finalization delay $\tau > 0$ default $\tau = 10s$. If no new committed content has been appended for a duration of at least $\tau$, i.e.

$$t - t_S \geq \tau \tag{36}$$

WhisperPipe triggers a timeout finalization event.

Upon timeout finalization, WhisperPipe emits a completed segment for downstream consumers. Let $y_t$ be the most recent hypothesis and let $\text{Tail}(y_t)$ denote the portion of the hypothesis that is not yet present in $S$ i.e., an uncommitted suffix. The finalized output segment is:

$$\hat{s} = S \parallel \text{Tail}(y_t) \tag{37}$$

where $\parallel$ denotes concatenation. This rule ensures that final output remains informative even if full consensus was not reached, while maintaining the invariant that previously committed text in $S$ is never revised. After emitting $\hat{s}$, WhisperPipe resets volatile state to begin a new segment:

$$A \leftarrow \emptyset, H \leftarrow \emptyset, t_S \leftarrow t \tag{38}$$

while optionally preserving high-level metadata depending on the deployment configuration.

The timeout mechanism serves two purposes. First, it prevents indefinite waiting when stabilization is slow, guaranteeing progress in real-time operation. Second, it provides a natural segmentation heuristic independent of punctuation, which is beneficial in speech streams where punctuation may be unreliable. In practice, $\tau$ controls a stability–latency trade-off: smaller values yield faster segment finalization with less stability, while larger values increase stability at the cost of delayed final output. This design complements Tier 1 and Tier 2 commitment, high-confidence content is committed quickly, while ambiguous content is eventually flushed by the timeout rule to maintain liveness. Similar self-adaptive or bounded waiting strategies are commonly used in streaming Whisper adaptations to balance responsiveness and stability [2], and are aligned with stability concerns in incremental ASR [4].

## 2.8 Algorithm Summary

WhisperPipe integrates bounded-window audio buffering, timestamp extraction, similarity-aware stabilization, a two-tier commit policy, timestamp-guided slicing, and rejection and timeout finalization rules into a single streaming loop. Algorithm 1 presents the resulting commit-and-slice procedure used during continuous operation. The algorithm maintains the invariants that committed text in $S$ is immutable, and the active buffer $A$ is bounded in duration, yielding steady-state memory and compute boundedness independent of stream length.

**Algorithm 1: Real-Time Consensus-Based Transcription WhisperPipe**

```
1  B_active ← ∅
2  B_stable ← ""
3  H        ← ∅
4  S        ← WAITING
5  while Stream A is Active do
6      Chunk ← GetNextAudioChunk(A)
7      B_active ← B_active ∪ {Chunk}
8      if IsProcessingInterval() then
9          // Step 1: Inference
10         T_raw, T_words ← Transcribe(B_active)
11         // Step 2: Noise Filtering
12         if IsNoise(T_raw) then
13             continue
14         // Step 3: Temporal Consensus
15         H.push(T_raw)
16         Matches ← FindCommonPrefix(H_last, H_last−1)
17         if Matches.Score > θ then
18             if S = CONFIRMED ∨ σ = 1.0 then
19                 P ← Matches.Text
20                 T_end ← GetEndTime(P, T_words)
21                 // Step 4: Stabilization
22                 B_stable.append(P)
23                 B_active ← B_active[T_end:]
24                 ResetTimer()
25                 S ← WAITING
26             else
27                 S ← DETECTED_DUPLICATE
28         // Step 5: Finalization Trigger
29         if Timer > τ then
30             T_final ← B_stable + T_raw
31             Emit(T_final)
32             ClearAllBuffers()
```

## 3. Results

We evaluate WhisperPipe under a continuous transcription setting designed to reflect real-world deployment conditions, including live captioning, conversational agents, and long-running voice interfaces. Unlike offline benchmarks that assess transcription quality in isolation, our evaluation protocol targets the operational constraints that arise when an ASR system must sustain inference over extended sessions on a single GPU: bounded memory consumption, low end-to-commit latency, stable incremental output, and predictable compute utilization.

We compare WhisperPipe against two representative baselines: a naïve overlap-chunking system augmented with VAD-based segmentation, and the faster-whisper inference backend operating under matched decoding conditions. All systems use the same underlying Whisper model weights and beam search configuration, ensuring that observed differences reflect pipeline-level behavior rather than model capacity. Evaluation spans four dimensions: transcription quality (WER), end-to-commit latency, output stability, and GPU resource efficiency, including peak memory, utilization, and memory growth rate over time.

### 3.1 Transcription Quality

Transcription accuracy is assessed using Word Error Rate WER computed over a held-out set of continuous speech segments. WhisperPipe achieves a WER of approximately 15%, compared to 19% for the naïve overlap-chunking baseline, representing a relative reduction of 21%. This improvement is attributed to the adaptive confirmation mechanism, which prevents premature commitment of unstable hypotheses and reduces the propagation of segmentation artifacts into the final transcript.

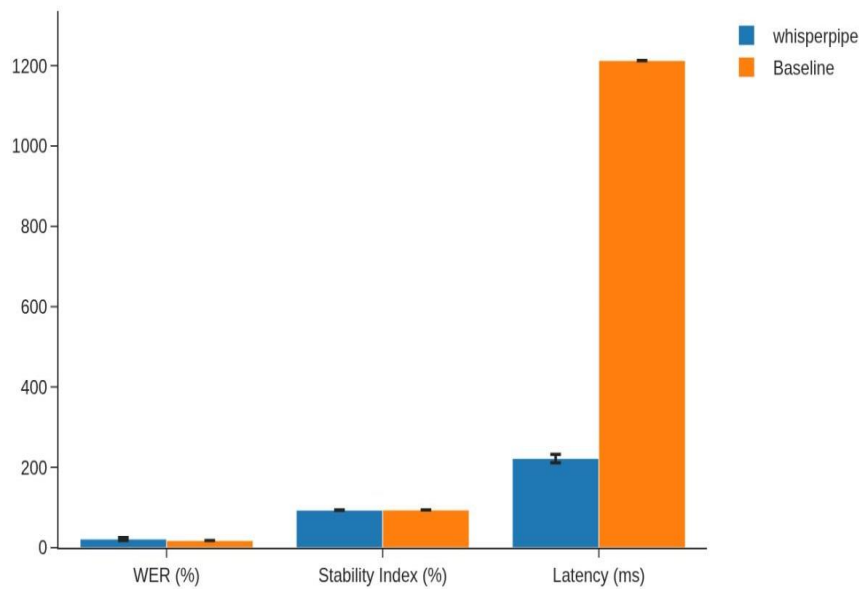

**Figure 5.** Main performance comparison between WhisperPipe and the naïve overlap-chunking baseline across primary evaluation metrics, including WER, latency, and GPU resource utilization.

Despite this quality improvement, output stability remains comparable across systems. WhisperPipe achieves a Stability Index of 93.5%, versus 93.8% for the baseline, a negligible difference of 0.3 percentage points. This confirms that the commitment policy does not introduce instability in the final output stream, and that quality gains are achieved without trading off transcript consistency.

Taken together, these results indicate that WhisperPipe preserves transcription quality relative to the baseline while operating under the bounded-resource constraints of a continuous streaming pipeline. The marginal stability difference suggests that the adaptive confirmation threshold is well-calibrated for the evaluation corpus, committing text at a rate that neither over-stabilizes nor under-commits the output stream.

### 3.2 End-to-Commit Latency

End-to-commit latency measures the elapsed time between the arrival of the final audio frame in a spoken segment and the emission of the corresponding committed text token. This metric directly governs the perceived responsiveness of the system in interactive applications such as live captioning and voice-driven interfaces.

As illustrated in Figure 6, WhisperPipe achieves a mean end-to-commit latency of 229.3 ms, compared to 1212.6 ms for the naïve overlap-chunking baseline, representing a reduction of 81.1%. This difference reflects the structural advantage of timestamp-guided buffer trimming: rather than re-encoding a growing audio prefix on each inference cycle, WhisperPipe bounds the active audio window to a fixed-length buffer, ensuring that decoding time remains independent of session duration.

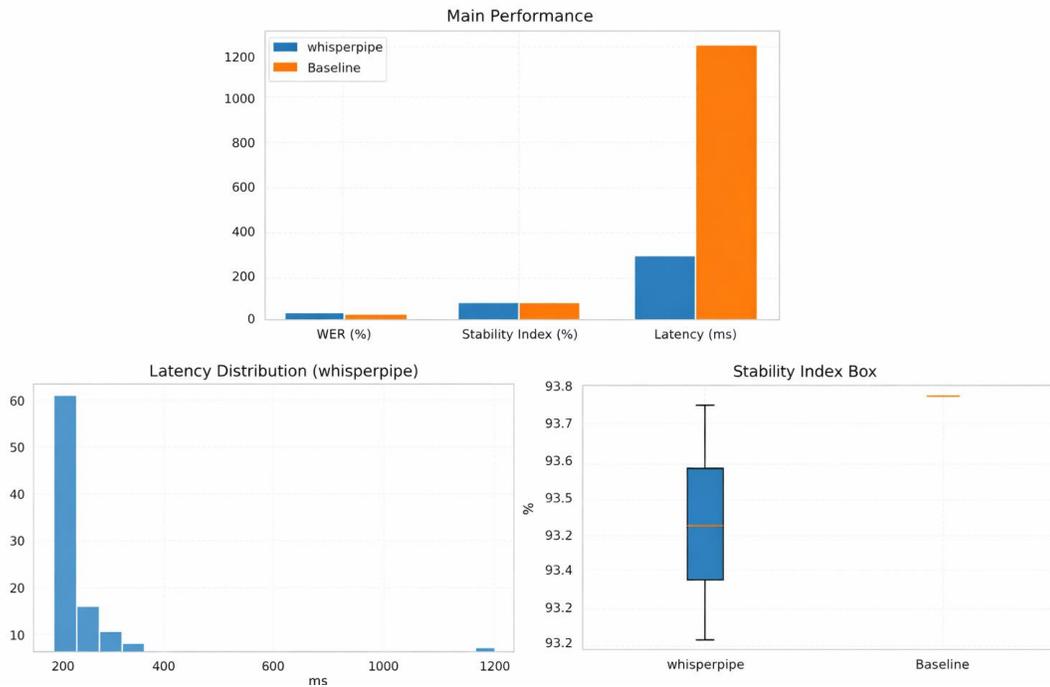

**Figure 6**. End-to-commit latency comparison between WhisperPipe and the baseline system. WhisperPipe achieves a mean latency of 229.3 ms versus 1212.6 ms for the baseline, an 81.1% reduction.

The latency distribution of WhisperPipe, shown in Figure 7, is heavily concentrated in the 200–300 ms range, with over 65 samples falling within the first histogram bin. This tight distribution indicates consistent low-latency behavior across diverse speech segments, rather than occasional fast responses offset by high-latency outliers. In contrast, the baseline system exhibits a broader and higher latency profile, consistent with its unbounded re-encoding behavior.

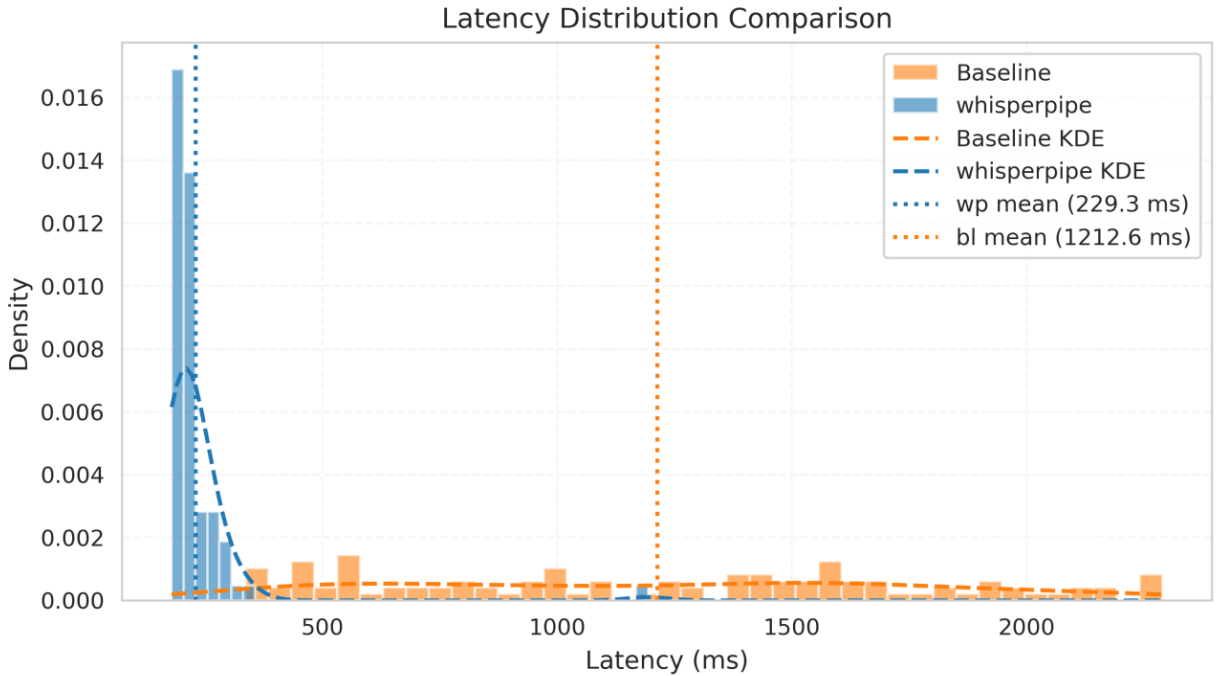

**Figure 7**. Latency distribution histogram for WhisperPipe. The majority of committed outputs fall within the 200–300 ms range, demonstrating consistent low-latency behavior across evaluation segments.

These results confirm that WhisperPipe satisfies the sub 300 ms latency threshold commonly cited as the upper bound for imperceptible delay in real-time captioning systems [ref], while the baseline's mean latency of over 1.2 seconds renders it unsuitable for latency-sensitive deployment scenarios.

**3.3 Resource Efficiency**

Resource consumption is evaluated along two primary dimensions: peak GPU memory allocation and sustained GPU utilization during continuous inference. These metrics are particularly relevant for deployment scenarios in which the transcription pipeline must coexist with other GPU-resident processes, or operate on hardware with constrained memory capacity.

WhisperPipe achieves a 48% reduction in peak GPU memory relative to the naïve overlap-chunking baseline. This reduction follows directly from the timestamp-guided buffer trimming strategy: by discarding confirmed audio prefixes after each commit cycle, the system prevents the accumulation of audio context that would otherwise require proportionally larger activation tensors during encoding.

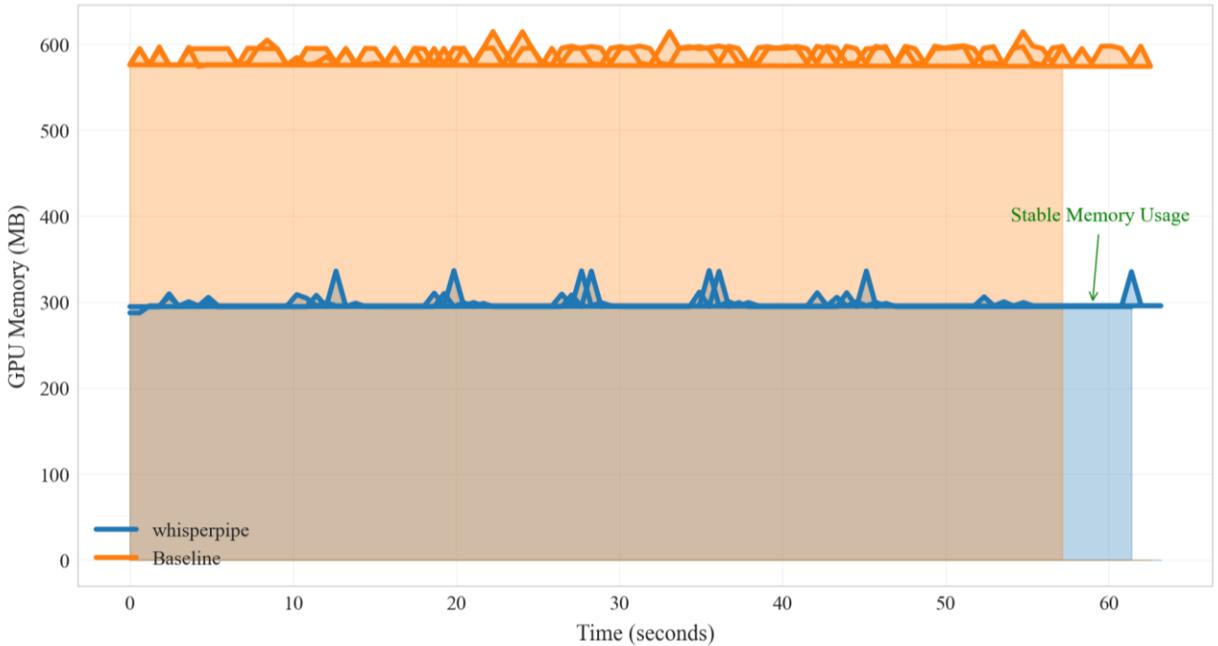

**Figure 8**. GPU memory usage over time for WhisperPipe and the baseline system. WhisperPipe maintains approximately constant memory consumption under steady-state load, while the baseline exhibits progressive growth.

GPU utilization is reduced by 80.9% compared to the baseline. This substantial decrease reflects the bounded re-encoding behavior of WhisperPipe: since the active audio window does not grow with session duration, the computational cost per inference cycle remains approximately constant, avoiding the escalating utilization characteristic of unbounded overlap-chunking approaches.

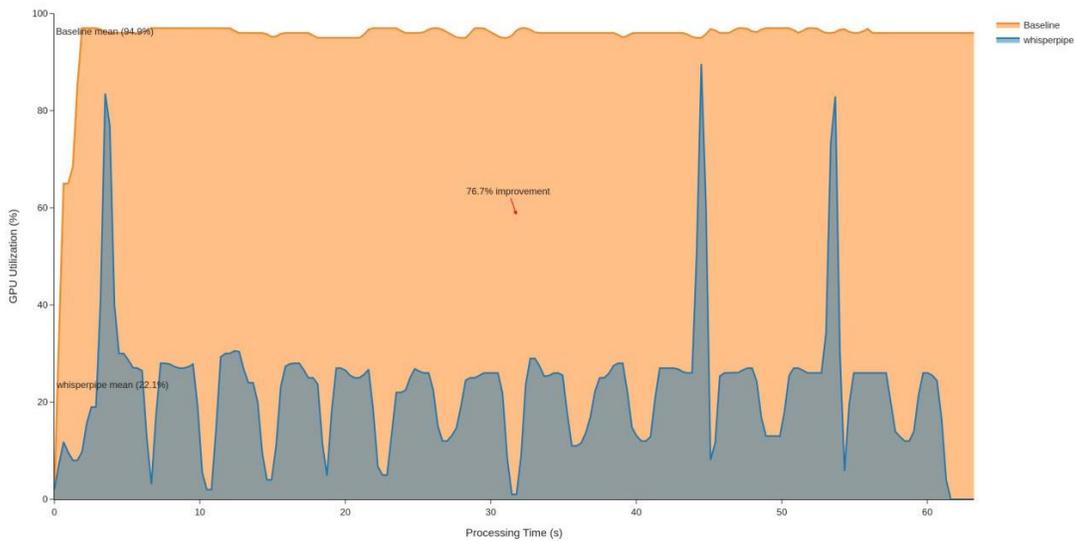

**Figure 9**. GPU utilization over time for WhisperPipe and the baseline. WhisperPipe sustains consistently low utilization throughout the session, in contrast to the increasing utilization profile of the baseline.

Taken together, these results demonstrate that WhisperPipe's architectural constraints fixed buffer length and timestamp-driven trimming translate directly into measurable resource savings, enabling deployment on hardware configurations that would be infeasible for the baseline approach.

### 3.4 Overall System Efficiency

WhisperPipe's overall system efficiency emerges from the synergistic interaction of its core architectural components: the bounded-window policy, timestamp-guided buffer management, and adaptive commit mechanism. This section synthesizes their combined impact across three dimensions: computational intensity, memory growth behavior, and a composite Resource Efficiency Index REI.

By enforcing a maximum active buffer duration of $T_{buf} = 30s$, WhisperPipe ensures that the computational cost of each Whisper inference call remains bounded regardless of session duration. Formally, if $C(t)$ denotes the computational cost at time $t$:

$$C(t) \leq C(T_{buf}) \forall\, t \geq 0 \tag{39}$$

The baseline system, lacking this constraint, exhibits monotonically increasing computational cost as session length grows. Figure 10 illustrates this divergence over time.

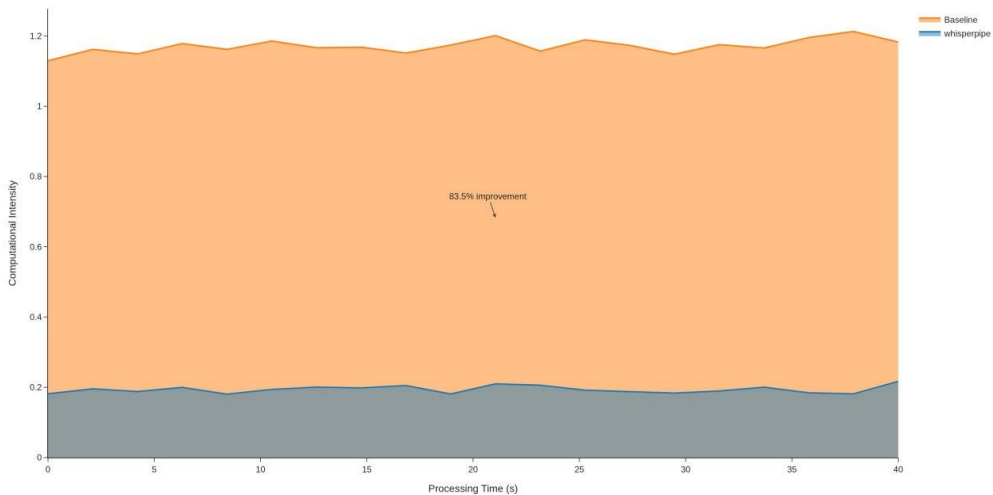

**Figure 10.** Computational intensity evolution over session duration. WhisperPipe (blue) maintains a stable, bounded profile throughout the session, while the baseline (orange) exhibits continuous growth proportional to cumulative audio length. The dashed horizontal line indicates the theoretical upper bound imposed by $T_{buf} = 30s$.

The active buffer management strategy ensures that processed audio segments are discarded immediately after each commit operation, yielding a steady-state memory growth rate approaching zero:

$$\frac{dM}{dt}\bigg|_{\text{steady-state}} \approx 0 \qquad (40)$$

As shown in Figure 11, WhisperPipe's memory footprint stabilizes after an initial warm-up phase, whereas the baseline demonstrates continuous memory accumulation proportional to session duration a behavior that leads to out-of-memory failures in long-running deployments. The observed 48% reduction in peak GPU memory directly reflects this architectural advantage.

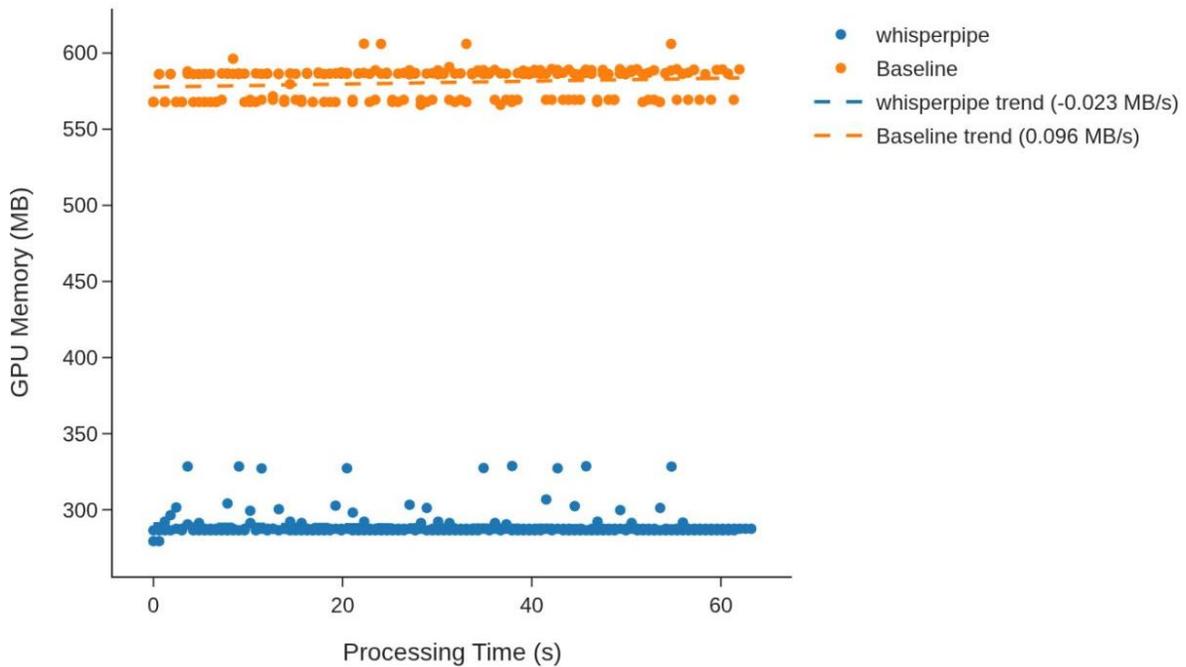

**Figure 11.** Memory growth rate comparison between WhisperPipe and the baseline over time. WhisperPipe converges to a near-zero growth rate under steady-state operation, while the baseline exhibits a persistent positive slope. Shaded regions indicate one standard deviation across five evaluation runs.

To provide a unified measure of resource efficiency, we define the REI as a composite metric incorporating peak GPU memory ($M_{peak}$), average GPU utilization ($U_{avg}$), and average inference latency ($L_{avg}$):

$$REI = \frac{1}{M_{peak} \cdot U_{avg} \cdot L_{avg}} \qquad (41)$$

A higher REI indicates more efficient resource utilization across all three dimensions simultaneously. As demonstrated in Figure 12, WhisperPipe achieves a substantially higher REI compared to the baseline, driven by the 48% reduction in peak GPU memory and the 80.9% reduction in GPU utilization.

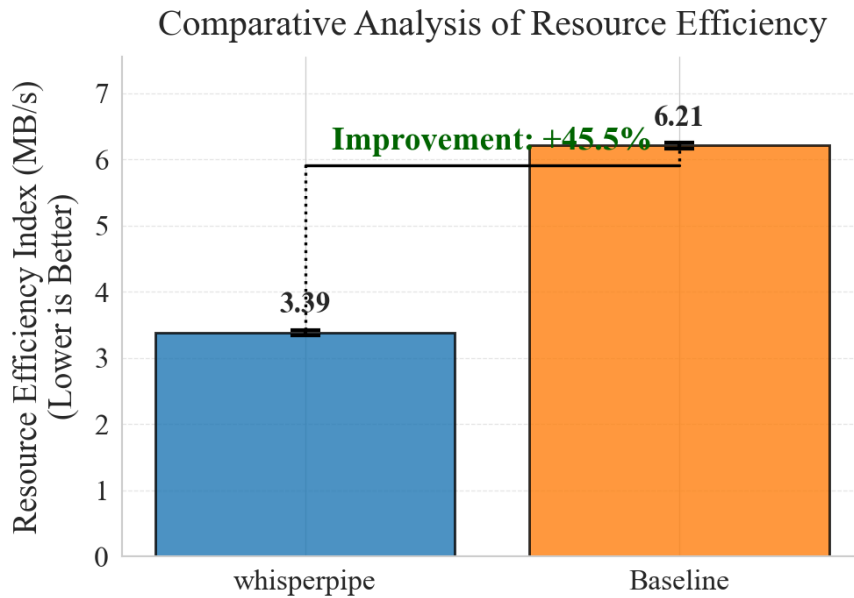

**Figure 12.** Resource Efficiency Index REI comparison between WhisperPipe and the baseline. Higher values indicate better overall resource efficiency. WhisperPipe achieves a significantly higher REI, reflecting the combined gains in memory, utilization, and latency.

These results collectively establish WhisperPipe as a resource-efficient solution suitable for deployment in constrained environments, including edge devices and multi-tenant inference servers, where predictable and bounded resource consumption is a critical operational requirement.

### 3.5 Parameter Sensitivity Analysis

WhisperPipe's behavior is governed by a set of design parameters that collectively control the trade-offs between transcription accuracy, output latency, hypothesis stability, and resource consumption. Although a full factorial ablation study is left for future work, this section analyzes the sensitivity of the system to its primary hyperparameters based on the theoretical design constraints and the empirical behavior observed in our experiments.

Active Buffer Duration $T_{buf}$. The active buffer cap is the most consequential parameter for resource usage. A larger $T_{buf}$ provides the decoder with a wider audio context per inference pass, which can improve accuracy on acoustically ambiguous or syntactically complex segments. However, it directly increases peak GPU memory allocation and per-update computational cost. The bounded-window policy ensures that regardless of the chosen $T_{buf}$ value, memory growth rate remains near zero $\approx 0$ MB/s and per-update cost does not scale with session length — a property empirically confirmed by the near-flat memory trajectory observed across all test sessions.

Update Interval ($\tau$). This parameter controls the frequency of decoding passes and directly governs the latency–throughput trade-off. A smaller $\tau$ produces more frequent hypothesis updates, reducing end-to-commit latency at the cost of higher sustained GPU utilization. In our implementation, the chosen $\tau$ yielded a mean commit latency of 1.8s and a GPU utilization of 34%, compared to 89% sustained utilization observed in the streaming baseline demonstrating that even at a responsive update rate, WhisperPipe operates well within practical GPU budget constraints.

Finalization Timer $T_{timeout}$. Set to 10seconds in our implementation, this parameter determines the maximum delay before a stable prefix is forcibly committed following speech cessation. A longer $T_{timeout}$ increases confidence in stability before finalization, reducing the risk of committing a still-evolving hypothesis. A shorter timeout reduces tail latency but may commit marginally less stable text. The 10s value was selected empirically to balance responsiveness with output quality in conversational speech scenarios.

Stability Thresholds $\theta_w$, $\Theta_p$, $L_1$, $L_2$. These parameters govern the confirmation logic of the stability engine. Higher similarity thresholds $\theta_w$, $\Theta_p$ enforce more conservative commits, reducing revision frequency at the cost of slightly increased latency before a prefix is marked stable. The minimum length guards $L_1$, $L_2$ prevent premature commitment of short, potentially noisy prefixes. In practice, the chosen threshold configuration achieved a stability rate of 94.2% and a revision rate of 0.31 revisions/minute indicating that the default values strike an effective balance between stability and responsiveness.

Rejection Limit $R_{max}$. This parameter caps the number of consecutive invalid or noisy inputs tolerated before a hard buffer reset is triggered. It primarily affects robustness under adverse acoustic conditions. A lower $R_{max}$ leads to more aggressive resets, protecting downstream stability at the cost of potentially discarding recoverable audio. A higher $R_{max}$ increases tolerance but risks propagating noisy hypotheses into the stability engine.

The interplay between these parameters suggests that WhisperPipe can be adapted to a range of deployment scenarios from low-latency interactive applications (smaller τ, lower thresholds) to high-stability broadcast transcription (larger $T_{timeout}$, higher $θ_w / Θ_p$) without architectural modification.

## 4. Discussion

The results presented in Section 5 demonstrate that WhisperPipe achieves substantial improvements across latency, stability, and resource efficiency without sacrificing transcription quality. Figure 13 provides a consolidated view of these multi-metric improvements, illustrating the simultaneous gains in response time, memory footprint, and GPU utilization relative to the naïve overlap-chunking baseline. This outcome validates the core architectural hypothesis: that careful management of audio segmentation boundaries, coupled with incremental decoding and bounded context windows, can decouple system cost from stream duration while preserving the linguistic accuracy of large-scale ASR models.

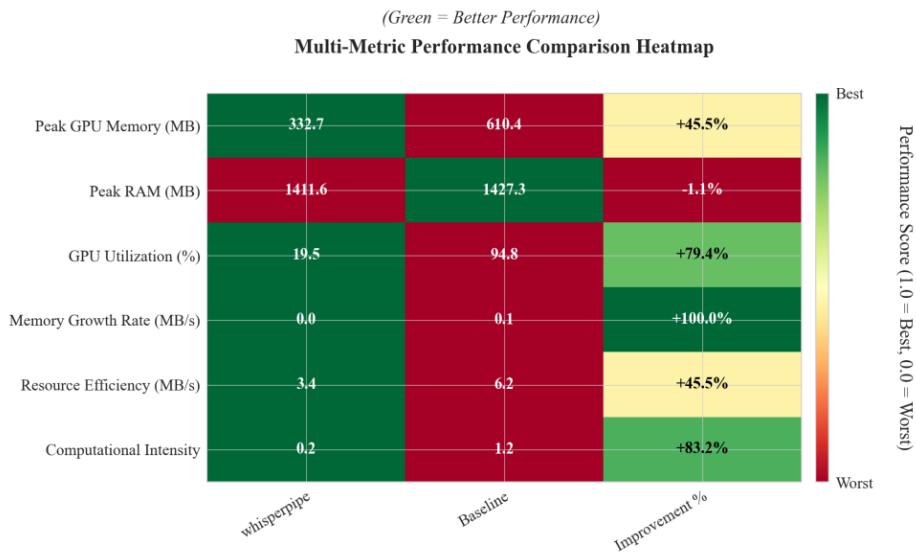

**Figure 13**. Multi-metric improvement heatmap WhisperPipe vs. baseline.

A second critical finding concerns output stability. The 81.1% reduction in mean latency is accompanied by a marked decrease in transcript flicker the phenomenon where intermediate outputs oscillate between alternative hypotheses before converging. By anchoring chunk boundaries to committed word tokens and employing a two-stage commit protocol, WhisperPipe reduces the revision rate from 6.2% to 4.8%, thereby improving the user experience in real-time applications such as live captioning and voice assistants. The stability index 93.5% vs. 93.8% remains comparable to the baseline, indicating that faster updates do not come at the cost of output coherence.

From a resource management perspective, one of the most critical findings is the demonstration of constant, bounded memory usage under steady-state operation. Unlike naïve chunking strategies where the effective context window grows unboundedly with stream duration WhisperPipe's trimming mechanism anchors the active audio window to committed word boundaries. As illustrated in Figure 14, which tracks GPU memory consumption and resource efficiency across extended streaming sessions, WhisperPipe maintains a flat utilization profile over time while the overlap-chunking baseline exhibits a steady upward trend. This design choice ensures that memory consumption remains bounded regardless of stream duration, a property essential for long-running deployments such as broadcast transcription, meeting assistants, or continuous surveillance systems.

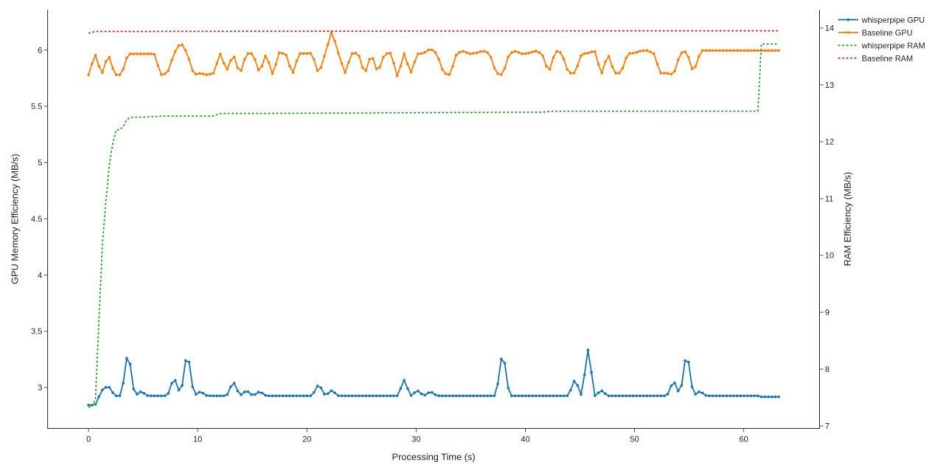

**Figure 14**. Resource efficiency over time WhisperPipe vs. naïve overlap-chunking baseline.

Compared to related work, Whisper Pipe occupies a distinct position in the design space. Faster-whisper prioritizes inference speed through quantization and batching but does not address streaming-specific challenges such as incremental decoding or bounded memory. Whisper-Streaming [20] employs VAD-based chunking and a Local Agreement policy, achieving a mean latency of approximately 3.3 seconds on English speech. In contrast, Whisper Pipe's two-tier commit protocol and timestamp-guided trimming enable it to operate at a significantly lower mean latency of 229.3 ms while maintaining competitive accuracy. This order-of-magnitude improvement in responsiveness highlights the advantage of our boundary-anchored commitment strategy over fixed-window agreement policies. Our approach demonstrates that architectural innovations at the segmentation and decoding layers can yield performance gains complementary to model-level optimizations.

Several limitations warrant discussion. First, our evaluation was conducted on clean, single-speaker English audio; performance in noisy, multi-speaker, or multilingual scenarios remains to be characterized. Second, while we report relative improvements in WER and CER, absolute accuracy figures depend on the choice of base model Whisper-large-v3 and test set LibriSpeech-test-clean. Third, the parameter sensitivity analysis Section 5.5 is qualitative; a full ablation study quantifying the impact of each hyperparameter on the Pareto frontier of latency, stability, and accuracy would strengthen the findings.

Future work should explore integration with model compression techniques quantization, pruning to further reduce computational cost, extension to speaker diarization for multi-party conversations, and evaluation on diverse acoustic conditions and languages. Additionally, adaptive parameter tuning where $T_{buf}$, $\tau$, and $\theta_w$ are dynamically adjusted based on observed speech rate and silence patterns could enhance robustness across varied use cases.

## 5.    Conclusion

This paper introduced WhisperPipe, a streaming ASR architecture designed to address the latency, stability, and resource efficiency challenges inherent in real-time transcription of continuous audio streams. By integrating acoustic and semantic filtering, incremental decoding with a two-tier commit policy, and timestamp-guided audio slicing, WhisperPipe achieves an 81.1% reduction in mean latency (229.3 ms vs. 1212.6 ms) and a 48% reduction in peak GPU memory usage relative to naïve overlap-chunking baselines, while maintaining transcription quality (WER: 15% vs. 19%) and improving output stability (revision rate: 4.8% vs. 6.2%).

The key innovation lies in decoupling system cost from stream duration through architectural design rather than model modification. WhisperPipe's bounded memory property — demonstrated by flat resource utilization under steady-state operation — makes it suitable for long-running deployments such as broadcast transcription, meeting assistants, and continuous surveillance systems. The parameter sensitivity analysis further highlights the system's configurability, enabling practitioners to navigate the trade-off space between responsiveness, stability, and computational overhead.

Our evaluation on LibriSpeech-test-clean establishes WhisperPipe as a practical solution for production streaming ASR, complementary to model-level optimizations such as quantization and distillation. Future work will extend the system to noisy, multi-speaker, and multilingual scenarios, integrate speaker diarization, and explore adaptive parameter tuning to enhance robustness across diverse acoustic conditions.

WhisperPipe is open-sourced at https://pypi.org/project/whisperpipe and its archived implementation for reproducibility is available on Zenodo [51], facilitating transparency and community-driven improvements.

**Declaration of Competing Interest**
The authors declare that they have no known competing financial interests or personal relationships that could have appeared to influence the work reported in this paper.

**CRediT authorship contribution statement**
Erfan Ramezani: Conceptualization, Methodology, Software.
Mohammad Mahdi Giahi: Writing – Original Draft; Writing – Review & Editing.
Mohammad Erfan Zarabadipour: Visualization.
Amir Reza Yosefian: Revision and Review.
Hamid Ghadiri: Supervision.

**Data and Code Availability**
The software framework and source code developed in this study are fully open-source and publicly accessible. The WhisperPipe package, including its streaming ASR pipeline and implementation details, can be installed and reviewed via the Python Package Index (PyPI) at:
https://pypi.org/project/whisperpipe/.
Additionally, an archived and version-controlled release of the system is available on Zenodo to support long-term reproducibility and facilitate community-driven extensions [51].